\pdfoutput=1

\documentclass[11pt]{article}

\usepackage[final]{acl}
\usepackage{times}
\usepackage{latexsym}
\usepackage{tcolorbox}
\usepackage{subcaption}
\usepackage{enumitem} 
\usepackage{amssymb}
\usepackage{csquotes}
\usepackage{booktabs}
\usepackage{makecell}
\usepackage[T1]{fontenc}

\usepackage[utf8]{inputenc}

\usepackage{microtype}

\usepackage{inconsolata}

\usepackage{graphicx}
\usepackage{etoolbox}
\usepackage{comment}
\usepackage{booktabs}
\usepackage{multirow}
\usepackage{fontawesome5}

\usepackage[normalem]{ulem}
\useunder{\uline}{\ul}{}

\title{\raisebox{-0.19\totalheight}{\includegraphics[height=6mm]{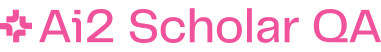}}: Organized Literature Synthesis with Attribution}

\author{Amanpreet Singh\thanks{\enspace Core contributors} \quad 
        Joseph Chee Chang$^*$ \quad 
        Chloe Anastasiades$^*$ \quad 
        Dany Haddad$^*$ \quad \\
        {\bf
        Aakanksha Naik \enskip
        Amber Tanaka \enskip
        Angele Zamarron \enskip
        Cecile Nguyen \enskip
        Jena D. Hwang \enskip 
        } \\
        {\bf
        Jason Dunkleberger \enskip
        Matt Latzke \enskip
        Smita Rao\enskip
        Jaron Lochner \enskip
        Rob Evans \enskip
        } \\
        {\bf
        Rodney Kinney \quad
        Daniel S. Weld \quad
        Doug Downey$^{*}$ \quad
        Sergey Feldman$^{*}$
        } 
        \\ [1mm]
        Allen Institute for AI
        \\
        {\tt\small \{amanpreets, sergey\}@allenai.org}
}

\begin{document}
\maketitle
\begin{abstract}
Retrieval-augmented generation is increasingly effective in answering scientific questions from literature, but many state-of-the-art systems are expensive and closed-source.  We introduce Ai2 Scholar QA, a free online scientific question answering application.  To facilitate research, we make our entire pipeline public: as a customizable open-source Python package\footnote{We use closed state-of-the-art LLMs.} and interactive web app, along with paper indexes accessible through public APIs and downloadable datasets.  We describe our system in detail and present experiments analyzing its key design decisions.  In an evaluation on a recent scientific QA benchmark, we find that Ai2 Scholar QA outperforms competing systems.

\begin{tabular}{lll}
\includegraphics[width=0.4cm]{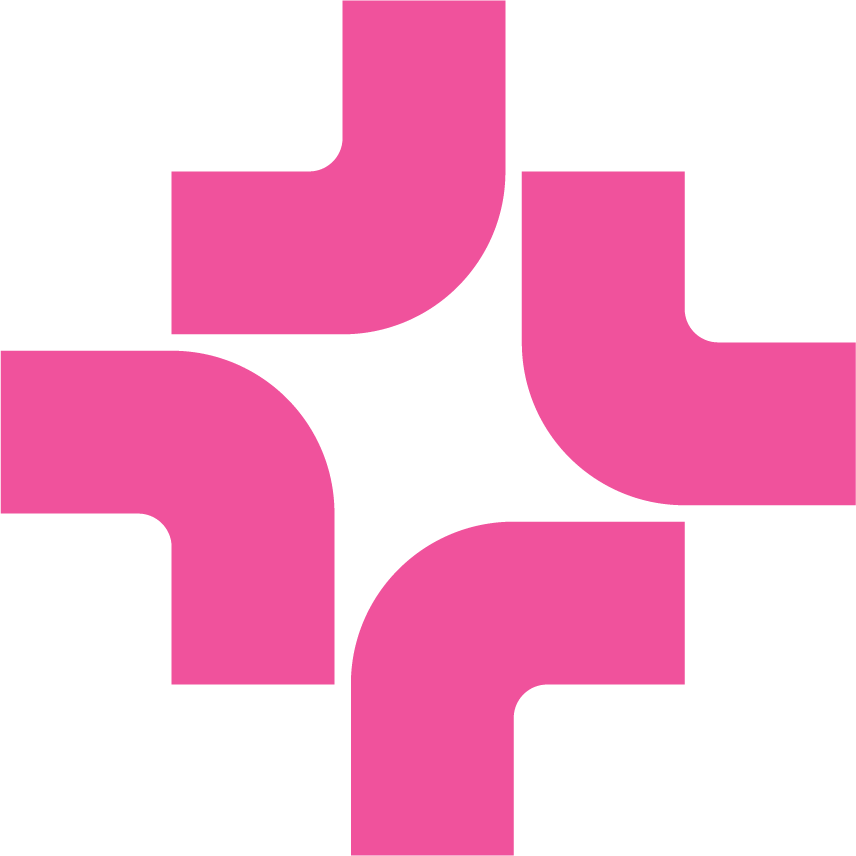}& \href{https://scholarqa.allen.ai/}{\footnotesize \texttt{qa.allen.ai}}\\
  \faGithub  & \href{https://github.com/allenai/ai2-scholarqa-lib/}{\footnotesize \texttt{allenai/ai2-scholarqa-lib}}\\
  \faYoutube  & \href{https://youtu.be/augQU982aGQ}{\footnotesize \texttt{Demo Video}}\\
  \faPython  & \href{https://pypi.org/project/ai2-scholar-qa/}{\footnotesize \texttt{Python Package}}
  
\end{tabular}
\end{abstract}

\section{Introduction}
Long-form scientific question answering systems use retrieval-augmented generation (RAG) \cite{Lewis2020RetrievalAugmentedGF} over scientific literature to answer complex questions.  These systems produce responses 
 that bring together relevant insights from dozens of papers to help users rapidly learn about a body of scientific work. Examples are OpenScholar \cite{Asai2024OpenScholarSS}, Elicit, 
Consensus, 
and others \S\ref{sec:related_work}. 

Most of these systems are expensive to use and closed source, relying on models, workflows, and retrieval solutions not shared publicly.  These issues create barriers for researchers who wish to study or build on the work.  In response, we introduce Ai2 Scholar QA, a free-to-use scientific QA system (\url{qa.allen.ai}), and share our key components as open source software and public APIs. 

\begin{figure*}
\centering
\includegraphics[width=1\linewidth]{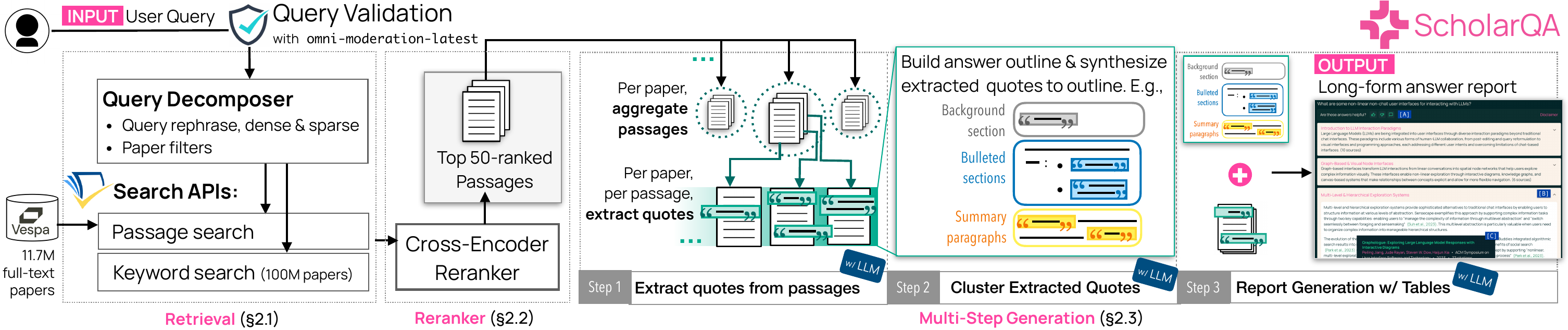} 
\caption{Scholar QA Pipeline Overview }
\label{fig:pipeline}
\vspace{-1.25em} 
\end{figure*}

Scholar QA follows a multi-stage pipeline (\autoref{fig:pipeline}) that starts by querying paper indexes: one from Semantic Scholar with over 100M abstracts, and a new index that we introduce in this work containing 11.7M full-text scientific papers.  The pipeline then re-ranks the retrieved passages with a cross-encoder, and finally prompts a Large Language Model (LLM) to filter, cluster, and synthesize the passages into an answer.  The final answer is presented to the user in a report with expandable sections of prose, bulleted lists, and tables. Claims in the answer are supported by citations, which can be clicked to reveal the cited paper's title and authors (with links to their corresponding Semantic Scholar pages), and in many cases relevant excerpt(s) from the paper, allowing for quick verification of the claim.

The system is based on open source code, enabling the community to reproduce and build on it.  We release the code for our pipeline, prompting workflow and Web application.  The retrieval indexes, including the new full text search index, are available as Semantic Scholar APIs and dataset downloads, and are continually updated with new articles \cite{Kinney2023TheSS}.  
Together, these resources can be combined with any generative LLM API to power a complete long-form scientific QA application. Our production system currently uses Anthropic's Claude 3.7 \citep{TheC3}.

We present analyses that justify key design decisions in our architecture in \S\ref{sec:evaluation}. Our choice of retrieval models and configuration is informed by evaluation over a collection of real and synthetic user queries and accompanying passages judged for relevance by a LLM, both of which we release publicly. We compare Scholar QA's answers against several baselines, demonstrating that it achieves state-of-the-art performance on the ScholarQA-CS benchmark \citep{Asai2024OpenScholarSS}.
Finally, we discuss the reception of Scholar QA by users.  The strong majority (85\%) of user feedback is positive, and the reported issues suggest important improvements for future work.
\section{Pipeline}
\vspace{-4pt}
The Scholar QA architecture (Figure~\ref{fig:pipeline}) has three primary components: 1) retrieval to identify relevant passages from a corpus of scientific literature; 2) a neural cross-encoder that re-ranks the passages to select the most relevant top-k; and 3) multi-step LLM generation to process the passages into a comprehensive report. Next, we describe each component of the pipeline in detail. 

\vspace{1mm}
\noindent \textbf{Query Validation.}
Prior to processing a query, we employ OpenAI's {\tt omni-moderation-latest}\footnote{\url{https://platform.openai.com/docs/guides/moderation}} model for safeguarding against potentially harmful content and return appropriate error messages. 
\subsection{Retrieval}
We use the Semantic Scholar API \citep{Kinney2023TheSS} for retrieval, specifically its endpoint for keyword search over paper abstracts, and our new endpoint for querying snippets from open-access papers. A query decomposer re-formulates the user query for each endpoint and retrieves up to 256 snippets and 20 abstracts. These texts are referred to as "passages" below.

\vspace{1mm}
\noindent \textbf{Query Decomposer.}
The two retrieval endpoints differ in their effective query formats (one targets keyword and the other semantic queries) and filtering of results based on the user's preferences for paper metadata (paper year, venue, field of study). In our query decomposition step, an LLM is prompted to re-format the user query into paraphrases appropriate for each endpoint, and to extract the user's requested settings for the metadata filters.  We use the outputs of this step for retrieval.

\vspace{1mm}
\noindent \textbf{Search APIs.}
\label{sec:fts_index}
The Semantic Scholar keyword search API is described in \citet{Kinney2023TheSS}. We introduce a new \href{https://api.semanticscholar.org/api-docs/#tag/Snippet-Text/operation/get_snippet_search}{/snippet/search endpoint}, which searches over a corpus of passages extracted from S2ORC \citep{Lo2020S2ORCTS}, loaded into a \href{https://docs.vespa.ai/}{Vespa} cluster with papers and passages. Papers  include metadata for filtering. Passages are derived from a paper's title, abstract, or body and can be filtered at the paper level. The index includes 11.7M full-text papers across the fields of study listed \href{https://api.semanticscholar.org/api-docs/#tag/Snippet-Text/operation/get_snippet_search}{here}, and a total of 285.6M passages.

Each passage is limited to 480 tokens and truncated at sentence and section boundaries where possible, having an overlap of one sentence (up to 64 tokens) with the preceding and following passages. Passage text is embedded with {\tt mxbai-embed-large-v1} \citep{emb2024mxbai} with binary quantization, and placed into a dense (approximate nearest neighbor) index, as well as a traditional sparse keyword index.

We first retrieve a union of embedding and keyword-based matches, applying any specified filters. The filtered results are ranked with a weighted sum of embedding similarity and bm25 scores.
\subsection{Reranking}
The passages obtained from the retrieval step are subsequently passed to a neural re-ranker and the top 50 results are retained. The re-ranker is a cross-encoder that encodes both the query and a candidate document simultaneously and outputs a relevance score used to rank the documents. We selected {\tt mxbai-rerank-large-v1} \citep{rerank2024mxbai} based on the results in \S\ref{sec:reranker_eval} and host it on \href{www.modal.com}{Modal} with a single NVIDIA L40S GPU. 
\subsection{Multi-step Generation}
The generation phase employs a three-step approach: first, the retrieved passages are processed to extract more precise quotes relevant to the query; second, the quotes are thematically clustered into separate sections appropriate for the answer; finally, a controlled generation process composes the final report one section at a time, synthesizing the quotes assigned to that section.

\vspace{1mm}
\noindent \textbf{Quote extraction.}
Passages from the retrieval stage can be lengthy and may contain extraneous information not useful for answering the user query \citep{Asai2023SelfRAGLT}. The quote extraction stage aims to select only the most relevant quotes from the passages to improve the precision of the answer.

We instruct an LLM to extract verbatim quotes 
that directly contribute to answering the query \citep{Slobodkin2024AttributeFT}.  As input to the extraction, we gather all passages from the re-ranker for a given paper, and concatenate these to the abstract of the paper. This aggregation helps create a richer context conducive to extracting relevant quotes. The LLM processes each paper's content independently and returns the selected quotes separated by ellipses. If the entire paper context is deemed irrelevant, it is discarded from further processing.

\vspace{1mm}
\noindent \textbf{Answer Outline and Clustering.}
\label{sec:outline}
For generating a comprehensive research report, the effective organization of reference materials is essential 
for its overall coherence. We propose a thematic outline framework where the answer is divided into sections representing topics, and the reference quotes are assigned to these topics. This mapping allows the system to selectively focus only on the pertinent subset of quotes when synthesizing a section. 

First, the LLM is instructed to generate a list of themes in logical order and the appropriate synthesis format for each theme, independent of the quotes from the previous step. The first section is always an introduction or background to provide the user the basics 
for understanding
the answer. The format of each section can be either a paragraph or a bulleted list, serving different information needs. Paragraphs convey nuanced summaries from multiple papers, while bulleted lists enumerate related papers (e.g., models, datasets, or interactive systems). 
These list are also the catalyst for generating the comparison tables (see  \S\ref{para:table_gen}). Following this, the sections are assigned 0 or more quotes. In case no quote is assigned to a section, it is generated completely from the LLM weights.

\vspace{1mm}
\noindent \textbf{Report Generation.}
With the answer outline in place, each section of the report is synthesized serially conditioned on the query, reference sources, and the sections prior to it. The LLM is also instructed to generate a TLDR for each section. The references are either the quotes assigned to the section or abstracts of papers that are cited within these quotes. This citation following method allows the LLM to condition on and cite foundational sources which are not uncovered in retrieval. 
The LLM is instructed to cite the sources for each claim in the generated section text and cite generations from its parameters as {\tt LLM Memory}.

\vspace{1mm}
\noindent \textbf{Paper Comparison Table Generation.}
\label{para:table_gen}
Since bulleted list sections typically include closely related papers (e.g., different \emph{datasets}), we additionally generate tables that compare and contrast all papers cited in that section using common aspects (e.g., \emph{size} and \emph{annotation method}). This pipeline is detailed in \citet{Newman2024ArxivDIGESTablesSS}. At a high level, the inputs are the query to Scholar QA, the section title, and the abstracts of all papers cited in the section. An LLM first produces a set of common aspects (columns) to compare papers (rows). Each cell (paper-aspect pair) is filled with a value using the full-text of the paper. Finally, as not all aspects are applicable to every paper (e.g., one paper might not be about a dataset), we filter out columns and rows with a high proportion of missing values. \autoref{fig:table} [A] shows an expanded table in Scholar QA where related papers from a section are compared across a set of common aspects ([B]).

\section{Scholar QA: Interface and Source Code}

Scholar QA is open-sourced as an extensible Python package ({\tt ai2-scholar-qa}) and a Typescript and React-based interactive web application. The LLM functionality of Scholar QA is implemented with {\tt litellm}, which supports swapping a variety of models using your own keys.
Thus, the community can build upon Scholar QA and easily visualize the results (examples in \autoref{sec:app_lib}). 
Below we describe the user experience of the demo.\footnote{Our production system has a few additional features like downloadable reports, login and links to other Ai2 systems.} 
%



\begin{figure*}[t!]
    \centering
    {%
\setlength{\fboxsep}{0pt}%
\setlength{\fboxrule}{0.1pt}%
    \fbox{\includegraphics[width=0.9\linewidth]{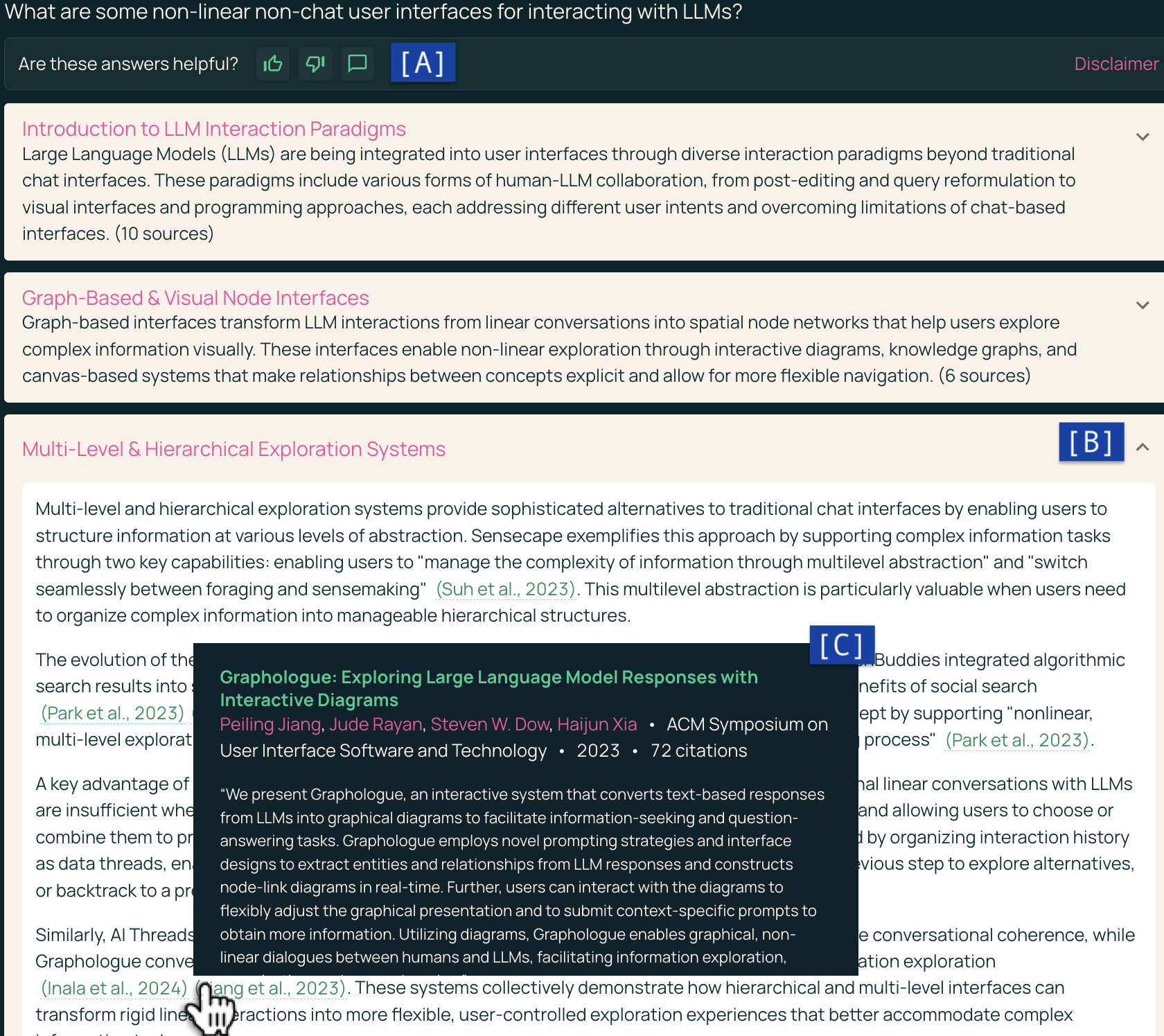}}
    }
    \caption{Multi-section [B] report generated by Scholar QA. References are linked to supporting excerpts [C]. Thumbs and free text feedback are collected for the full report [A], and also for each section and inline table.}
    \label{fig:sections}
\end{figure*}

\vspace{1mm}

\paragraph{Progress and Section Streaming.}
High system latency can hinder usability. On average, Scholar QA produces a full report in 2.5 minutes (N=500, $\sigma$=70s), which is comparable to modern LLM-based research tools. To further improve usability, the following designs were used: 1) Displaying detailed real-time progress of the system \cite{nielsen1994enhancing} so users can examine the number of papers, passages, and sections being processed. 2) Presenting each section as soon as it is generated, so users can begin browsing the first section in 50 seconds (N=500, $\sigma$=24s) post issuing a query (\autoref{sec: app_progress}).
\paragraph{Expandable Sections.}
By default, sections are collapsed showing only their titles, TLDR summaries, and number of cited sources. This gives users a gist of the information included in the report (Figure~\ref{fig:sections} [A]). Users can then click on the title of a section they wish to read to expand it ([B]).

\begin{figure}
    \centering
    {%
\setlength{\fboxsep}{0pt}%
\setlength{\fboxrule}{0.1pt}%
    \fbox{\includegraphics[width=1\linewidth]{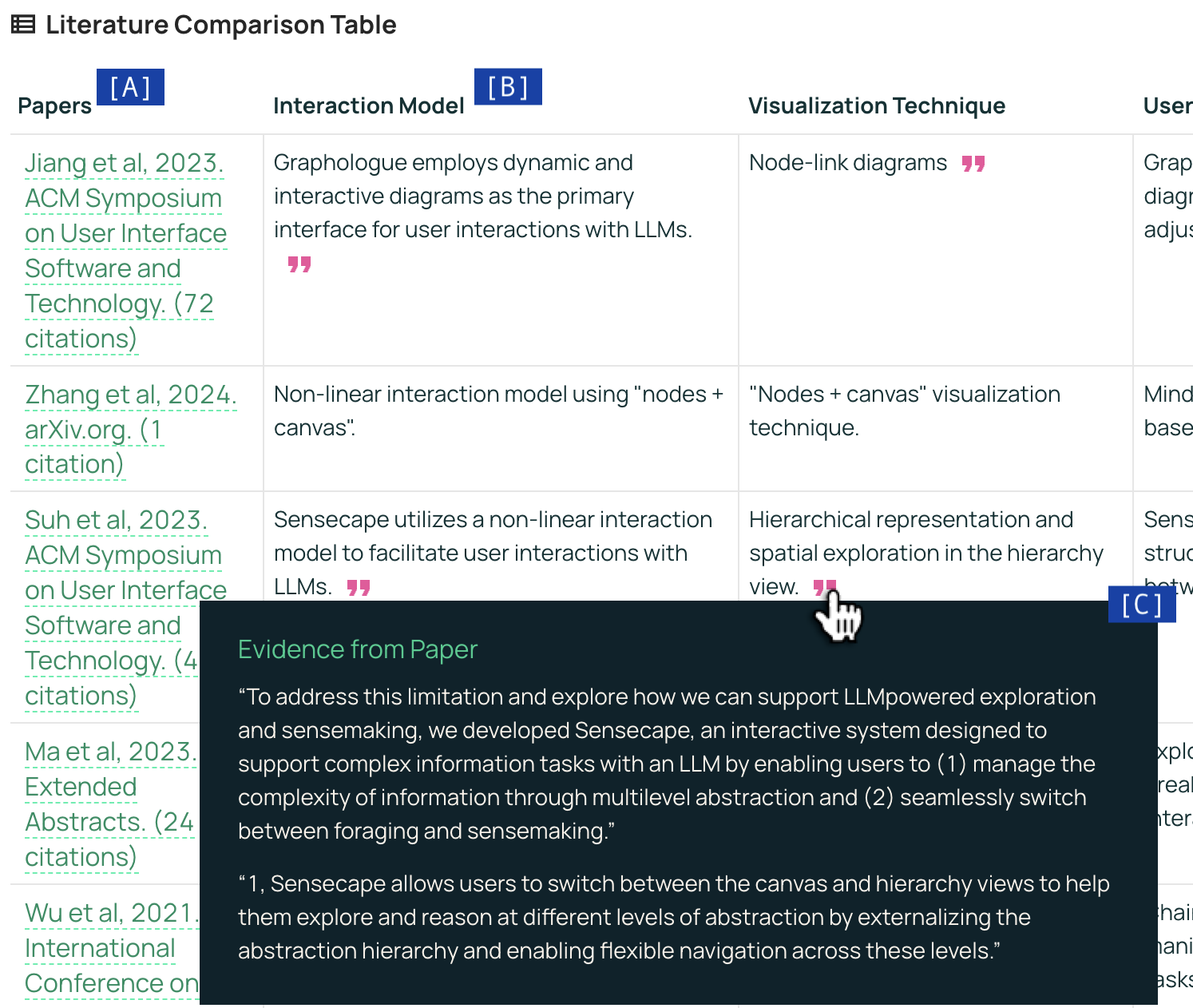}}
    }
    \caption{Inline tables compare papers [A] with common aspects [B] with values linked to supporting excerpts from the papers [C].}
    \label{fig:table}
    \vspace{-0.75em}
\end{figure}

\vspace{-10 pt}
\paragraph{References and Evidence Excerpts.}
To verify the claims in the report, users can click on the inline citations (Figure~\ref{fig:sections} [C]) or the pink excerpt icon in the inline table cells (Figure~\ref{fig:table} [C]) to bring up a popup paper card. From the paper card, they can see the relevant excerpts used during the generation or click on the title to open the paper directly.

\vspace{1mm}
\noindent \textbf{User Feedback Collection.}
We collect thumbs up/down or textual feedback for the whole report (Figure~\ref{fig:sections} [A]) and at each section and inline table. 

\vspace{1mm}
\section{Evaluation}
\label{sec:evaluation}
\subsection{Retrieval}
We tuned our retrieval setup by optimizing ranking over a dev set of 500 synthetic queries (see \autoref{sec:app_retrieval_queries}) and the top 1000 passages for each based on GIST embedding distance \citep{Solatorio2024GISTEmbedGI}. We generated binary relevance labels with \texttt{gpt-4-turbo} (see \autoref{sec:app_rel_prompt} for the prompt), which were found to have 80\% agreement with human annotators on a sample of 100 queries.


\vspace{1mm}
\noindent \textbf{Pipeline Tuning.}
We optimized several aspects of retrieval over this dev set: embedding model selection and quantization method for it, the components and weights in the final ensemble, and (when relevant) the target Matryoshka dimension for the embeddings \cite{kusupati2024matryoshkarepresentationlearning}.

\begin{figure}[!t]
    \centering
    \includegraphics[width=1.0\linewidth]{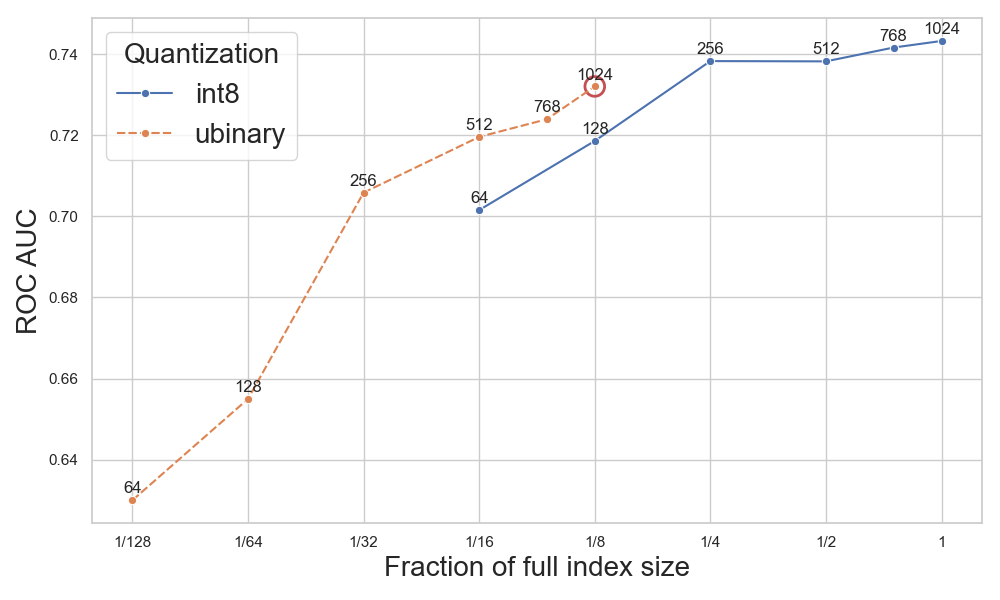}
    \caption{Embedding ranking performance for various compression methods and matryoshka cutoffs. The x-axis indicates the size of the vector index based relative to using \texttt{int8} quantization and the full embedding size. The red circle indicates the selected configuration. Embedding size is notated next to each point.}
    \label{fig:ranking}
    \vspace{-1em}
\end{figure}

We experimented with medium sized embedding models  based on top performers on the retriever and ranking tasks of the MTEB \citep{Muennighoff2022MTEBMT} leaderboard on HuggingFace. \autoref{tab:embedding_models} in \autoref{sec:app_embedding_models} lists our candidate models.
The \texttt{mxbai-embed-large-v1} \citep{emb2024mxbai} embeddings performed best over our dev set. Figure \ref{fig:ranking} validates our choice of quantization method and target Matryoshka dimension for these embeddings. We chose \texttt{ubinary} quantization with no Matryoshka truncation, (indicated by a red circle on the plot) since it satisfied our storage constraints without a large drop in performance. We experimented with ensembling \texttt{SparseEmbed} \cite{Kong2023}, embedding cosine similarity, BM25, and chose the latter two (weight split of $(0.6, 0.4)$ respectively)  based on the results (See  \autoref{sec:app_ensemble}). 
The BM25 scores are normalized with min-max scaling before computing the ensemble score.

\subsection{Reranking}
\label{sec:reranker_eval}
We chose the re-ranker based on evaluation over a mixture of real scientific questions from the Stack Exchange Computer Science, Math, and Statistics communities, real research queries written by the authors and their colleagues, and synthetic ones generated by fine-tuning GPT-4o-mini over questions from the ScholarQA-CS dataset \citep{Asai2024OpenScholarSS}. For a given query, passages are retrieved and then awarded a relevance score in the range 0-3 with GPT-4o. We experiment with multiple state-of-the-art re-rankers \citep{chen2024bge,rerank2024mxbai,Asai2024OpenScholarSS}, and, as shown in \autoref{tab:sqa_eval}, {\tt mxbai-rerank-large-v1} gives the best results across the board (even outperforming its v2 model on our task). To reduce latency for deployment, we implemented optimizations like Pytorch model compilation. 
We release the evaluation data consisting of 2,426 queries and 225,618 passages.

\subsection{Generation}
\label{sec:gen_eval}
We evaluate the final output of Scholar QA on the ScholarQA-CS dataset which consists of expert-annotated rubrics for 100 Computer Science research questions. The question-specific expert rubrics account for 60\% of the final score, while the rest is computed based on global metrics of length, expertise and citations. We use GPT-4o \citep{Hurst2024GPT4oSC} as a judge with the utility provided by \citet{Asai2024OpenScholarSS} for automatic evaluation and compare against several baselines.


\begin{table}[]
\centering
{\fontsize{8.0}{10}\selectfont
\begin{tabular}{lccc}
\hline
Model (Size) & \makecell{Latency \\ (sec/query)} & \makecell{nDCG \\ @10} & mRR \\
\hline
bge-reranker-v2-m3 (568M)      & 0.14 & 0.913 & 0.973 \\
akariasai/ranker\_large (568M) & 0.14 & 0.906 & 0.970 \\
jina-reranker-v2-base (278M)   & \textbf{0.06} & 0.907 & 0.972 \\
mxbai-rerank-large-v1 (435M)   & 0.46 & \textbf{0.927} & \textbf{0.975} \\
mxbai-rerank-base-v1 (184M)    & 0.19 & 0.919 & 0.974 \\
mxbai-rerank-xsmall-v1 (70M)   & 0.11 & 0.911 & 0.970 \\
mxbai-rerank-base-v2 (0.5B)    & 0.40 & 0.918 & 0.974 \\
mxbai-rerank-large-v2 (1.5B)   & 0.70 & 0.911 & 0.975 \\
\hline
\end{tabular}
}
\caption{Cross encoder re-ranker results on our dataset of GPT-4o labels. The best results are \textbf{highlighted}.}
\label{tab:reranker_eval}
\vspace{-1em}
\end{table}

As shown in \autoref{tab:sqa_eval}, our system outperforms popular LLMs:
Llama 3.1 \citep{Dubey2024TheL3}, GPT 4.1 and Claude Sonnet 3.7 \citep{TheC3}. It even outperforms reasoning models such as Sonnet 3.7 Thinking \citep{Claude3S}, o1-mini \citep{OpenAIOS} and o3-mini \citep{ZhangOpenAIOS} overall on the Scholar QA-CS benchmark. This setup lacks any retrieval so the models generate the responses completely from parametric memory. The benchmark rewards attribution and supporting evidence as a measure of trust in the system, so these models score lower overall. The reasoning based models perform better than our system on the rubrics score, which suggests that they may be superior backbones for our system. However, due to the additional reasoning tokens, these models are more expensive and also significantly increase  latency. 

For contemporary QA systems, we compare against OpenScholar with GPT-4o\footnote{Our results are not identical to \citet{Asai2024OpenScholarSS} due to variance across LLM-as-a-judge runs. Their reported total score for OS-GPT-4o is 57.7. We re-ran the evaluation in order to obtain rubrics only scores, which they did not report.}, PaperQA2 \citep{Skarlinski2024LanguageAA}, Perplexity's {\tt Sonar Deep Research} and STORM \citep{shao-etal-2024-assisting}. PaperQA2 did not release their retrieval corpus, so we substitute it with our retrieval pipeline for a fair comparison. 
Scholar QA obtains the best scores both on rubrics and overall, with the variant using Claude 3.7 Sonnet as the backbone scoring 2.4 points higher than STORM. 
For these QA systems, we also evaluate the attribution quality based on ALCE \citep{Gao2023EnablingLL}, which proposes entailment between claims and evidence to compute citation precision and recall. Again, we use GPT-4o as a judge to predict entailment (See \autoref{sec:app_att_prompt} for the prompt) and treat each sentence in a response as a claim. Even with a report spanning multiple sections where all the sentences might not be cited, Scholar QA comes out far ahead of the other QA systems. Due to a lack of retrieval, this evaluation was not conducted when the LLMs are simply prompted to generate a response from memory. An interesting discovery from our analysis was that with an updated version of GPT-4o (i.e. {\tt gpt-4o-2024-11-20}) as the judge, the scores are inflated compared to using {\tt gpt-4o-2024-08-06}, even though the relative rankings are consistent (See \autoref{sec:app_gpt4o_updated_res}). For parity with \citet{Asai2023SelfRAGLT}, we report the rubrics and citation scores with the older and newer model as the judge, respectively. 

During our initial experiments, we restricted ScholarQA to only summarize the insights conditioned on the quotes extracted from retrieved passages. However, in cases where the retrieved passages were not relevant enough, the system failed to answer the question in favor of just discussing the information in the quotes. Moreover, for over 30\% of instances in ScholarQA-CS, the rubrics require background information, even though the question might not. So, we updated our system LLM prompts to -- a) Generate section text from memory if there is a lack of relevant retrieved passages and cite as LLM Memory and b) generate the first section as a background or introduction for the rest of the answer. The results reported here are obtained post these changes. 

To finalize the backbone LLM for the production web application we conducted an anonymized pairwise comparison among the authors of this work. We compare Claude 3.7 against 3.5. Out of 18 comparisons, Claude 3.7 Sonnet was the overwhelming favorite with 17 wins, reinforcing our hypothesis that (with no other changes) our system improves with newer and better backbone LLMs.

\begin{table}[!t]
\scriptsize
\centering
\setlength{\tabcolsep}{3pt}
\renewcommand{\arraystretch}{1}
\begin{tabular}{@{}lrr|lrrrl@{}}
\toprule
\multirow{2}{*}{Model}   & \multicolumn{2}{c|}{Score}  & \multirow{2}{*}{Model}  & \multicolumn{3}{c}{Score}                                 \\ \cmidrule(l){2-3} \cmidrule(l){5-7} 
 & Rubrics& Total   & & Rubrics& Total  & Cite  \\ \midrule
\multicolumn{3}{c|}{\emph{LLM Prompting (No Retrieval)}}     & \multicolumn{4}{c}{\emph{QA Systems}}               \\ \midrule

Llama 3.1-8B               & 48.8 & 47.3   &    
SQA-Claude 3.7 S                                              & {\ul 58.0}                             & {\ul \textbf{61.9}} & 48.1               \\

Llama 3.1-70B              & 52.4 & 48.6     &  
SQA-Claude 3.5 S & 52.6                 & 61.3     & \textbf{52.1}  \\


Claude 3.5 S       & 50.4 & 46.6     &
OS-GPT-4o     & 49.3        & 53.5    & 25.9  \\

Claude 3.7 S       & 61.5                      &    55.9 &   
PaperQA2              & 38.7   & 51.4  & 25.3  \\

\quad +Thinking  & 62.7                & 55.7 & 
Perplex. Sonar DR             & 38.7                 & 52.8   &  25.2  \\

GPT-4.1                 & {\ul\textbf{63.2}}                 & {\ul56.2}    &   STORM  & 54.2 & 59.5       & 40.2  \\
o1-mini                 & 62.3             &  55.5     \\
o3-mini                 & 60.6                & 50.2    &   \\ \bottomrule

\end{tabular}
\caption{Evaluation results on ScholarQA-CS benchmark. System responses are either generated by simply prompting LLMs with the questions or by issuing the queries to RAG based QA systems. Expert annotated rubrics only scores are reported in addition to the overall total. The overall best results are \textbf{highlighted} and best results within a category are {\ul underlined}. SQA: Ai2 Scholar QA, OS: Open Scholar, S: Sonnet, Claude 3.5 S: {\tt claude-3-5-sonnet-20241022}.}
\label{tab:sqa_eval}
\vspace{-1.5em}
\end{table}

\subsection{Real-world Usage and User Feedback}
\label{sec:feedback}
We have publicly deployed Scholar QA for 9 weeks, and received 30.2k questions from 8,219 unique visitors. On average, each response is about 2.4k words and costs \$0.50 to produce.  We observed 1,075 monthly repeated users who had issued queries on two distinct days over the course of a 30 day window. We analyze the user query types and the most prominent themes were \emph{deep-dive} into specific research topics (15k) and \emph{comparative analysis} of specific prior work (5k) (detailed distribution in \autoref{sec:app_query_types}).
A total of 2,433 thumbs feedback  were submitted (Figure~\ref{fig:sections} [A]) and 85\% were positive. These suggests real-world users benefited from using Scholar QA.

For insight into
the failure modes, we manually examined the 383 instances of  neutral/negative free-form feedback. Table \ref{tab:feedback_counts} lists the feedback types we identified along with their counts as of May 2025 (example feedback in \autoref{sec:app_feedback}). We hypothesize that follow-up questions may help address insufficient answer detail and cases with a lack of retrieved documents, while improved retrieval may help address incomplete or incorrect references and off-topic responses.
\begin{table}[!t]
\centering
\begin{tabular}{lr}
\toprule
Category & Count \\
\midrule
Incorrect or Missing References & 126 \\
Off-topic or Misunderstood Query & 113 \\
Request for More Detail or Specificity & 289 \\
General Feedback on Quality & 149 \\
Language or Format Issues & 78 \\
\bottomrule
\end{tabular}
\caption{Feedback Categories and Counts}
\label{tab:feedback_counts}
\vspace{-1em}
\end{table}

\section{Related Work}
\label{sec:related_work}
\paragraph{Scientific Question Answering.}
Answering scientific questions involves navigating scholarly sources and accurately retrieving and synthesizing them. Recently, OpenScholar~\citep{Asai2024OpenScholarSS} introduced a retrieval-augmented model designed explicitly for scientific literature synthesis with citation-supported responses with
significant improvement in accuracy and reduced citation hallucination. Scholar QA extends its capabilities by leveraging the latest state-of-the-art LLMs and an open source generation pipeline that filters literature into precise quotes and produces thematically organized and detailed answers.  STORM~\citep{shao2024assistingwritingwikipedialikearticles} synthesizes comprehensive, Wikipedia-like articles, a distinct task from long-form scientific question answering.  Other works have focused on literature review synthesis: LitLLM~\citep{Agarwal2024LitLLMsLF}, which like Scholar QA uses a structured planning-and-generation pipeline similar, and SurveyForge~\citep{Yan2025SurveyForgeOT}, which outlines heuristics before generation. Their code was not available at the time of our evaluation.
\citet{Zhou2025FromHT} present a survey categorizing AI-driven research support systems across various stages of the scientific process, including literature synthesis.




\paragraph{Commercial Tools for Scientific QA.}
Commercial RAG tools have emerged to facilitate research specifically tailored for scientific literature, such as Consensus \cite{consensus}, which synthesizes findings from research papers, Scite \cite{scite}, which evaluates claims by analyzing citation contexts, and Elicit \cite{elicit}, which supports structured scientific literature reviews. Other general-purpose tools also support scientific inquiries: Perplexity \cite{perplexity}, You.com \cite{you}, OpenAI Deep Research \cite{openai-deep-research} and Gemini Deep Research \cite{gemini-deep-research}.
Although these platforms leverage advanced retrieval and generation capabilities to facilitate literature reviews and deliver rapid insights, they can be too expensive for widespread academic use and typically lack transparency regarding their pipelines. In contrast, Scholar QA is free with open sourced code and access to search APIs that enable the research community to build upon it.
\section{Conclusion}

We present Ai2 Scholar QA, a freely-available long-form literature synthesis system that generates reports for complex scientific questions. We release key components as open source code and public APIs, and report experiments analyzing design decisions and demonstrate state-of-the-art results.

\section*{Limitations}
Supplementing the user feedback discussed in \autoref{sec:feedback}, we would like to outline some limitations of our system and evaluation and our plans to mitigate them as part of fuure work:
\begin{enumerate}[label=(\roman*), labelsep=0.5em]
    \item Ai2 Scholar QA uses proprietary  and closed-source LLM as the backbone for our production pipeline.  As shown in \autoref{tab:sqa_eval}, open source models lag behind the proprietary models in our evaluation. However, we are actively experimenting with open-sourced LLMs to replace the closed ones partially or completely in the pipeline.  The open-sourced models will be specifically trained to do well on long-form scientific question answering and each of the sub-tasks in our multi-step generation. Further, our code is open-sourced and can easily be used with potentially any available LLM api provider supported by litellm.
    \item We evaluate the answers generated by Scholar QA and compare against other systems on ScholarQA-CS dataset in \autoref{sec:gen_eval}. Even though the answer rubrics are collected via human annotation, the evaluation is only limited to questions in the Computer Science domain and further relies completely on an LLM as the evaluator. In ongoing work, we are investigating more accurate benchmarks for evaluating long form scientific answers. Our approach uses real queries posed by users to Scholar QA, and human preference labels over answers from multiple systems in not just Computer Science, but Biomedicine and other scientific domains. These labels can serve as not only for evaluation, but also as training signals for models.
    
\end{enumerate}
\section*{Acknowledgments}
We would like to thank the anonymous reviewers for helpful comments, suggestions and feedback on the manuscript. We would also
like to acknowledge the Ai2 ScholarQA users for providing constructive feedback that helped us improve the system. Finally, we thank David Albright for helping with the demo video, the Ai2 communications team for their help with user outreach, and Ai2 engineers and researchers for their help with user testing before launch.
\bibliography{0_main}
\clearpage
\appendix

\newpage
\section{Python Package Usage}
\label{sec:app_lib}
\begin{figure}[!h]
    \centering
    \fbox{
    \includegraphics[width=0.95\linewidth]{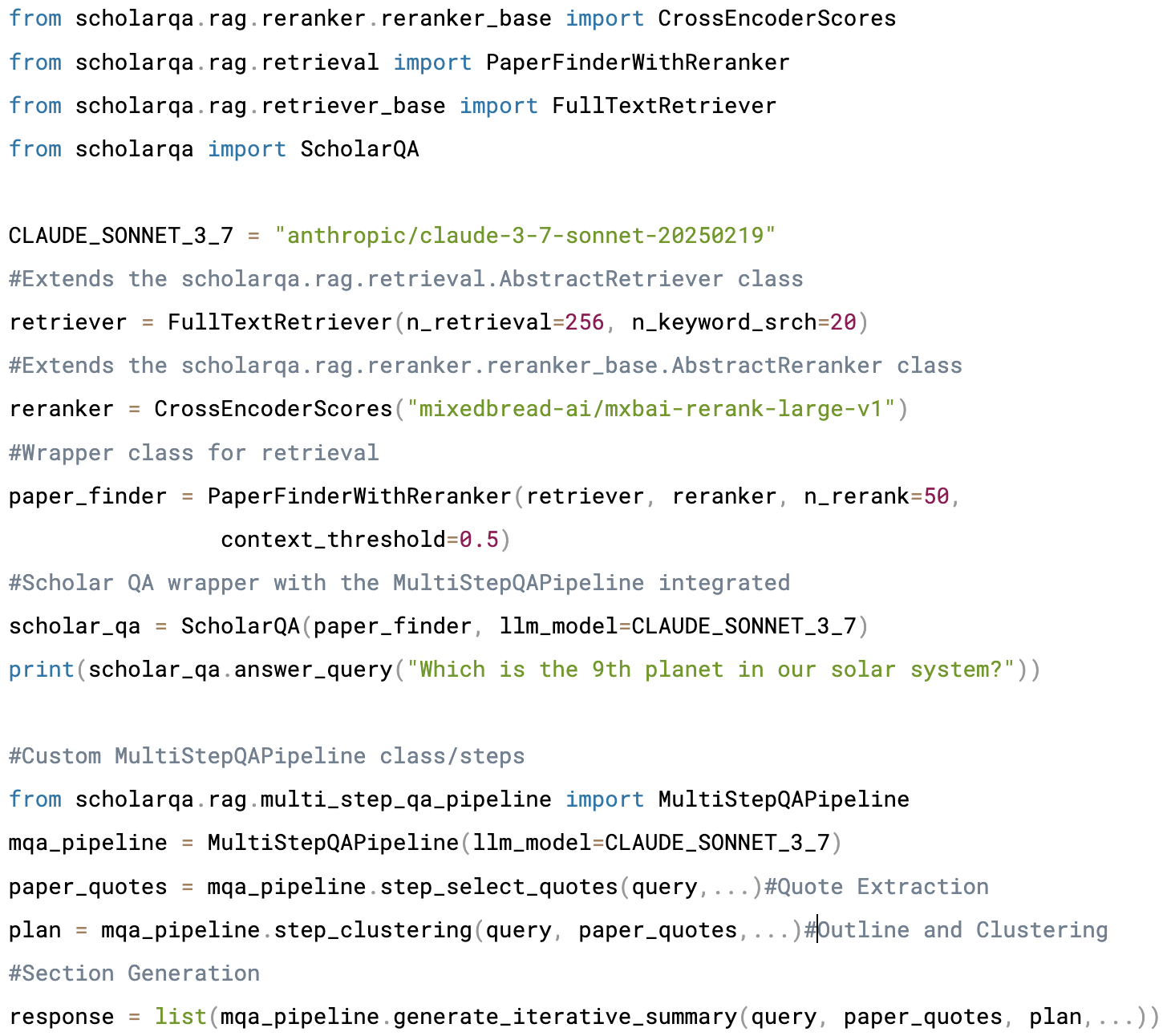}
    }
    \caption{{\tt ai2-scholar-qa} usage example}
    \label{fig:code_excerpt}
\end{figure}

\autoref{fig:code_excerpt} shows a minimal example of running the system pipeline with the ai2-scholar-qa python package and how every component can be extended or modified as the users see fit.

\section{Document Relevance Prompt}
\label{sec:app_rel_prompt}
We used the following prompt to obtain binary relevance labels, which agreed with human annotators 80\% of the time: 
\begin{displayquote}
\footnotesize
\texttt{If any part of the following text is relevant to the following question, then return 1, otherwise return 0. Non-english results are not relevant, results which are primarily tables are not relevant.}
\end{displayquote}

\section{Retrieval Tuning Query Generation}
\label{sec:app_retrieval_queries}
Queries for the dev set were obtained from three internal sources of human research questions, and a set of LLM generations. We experimented with several methods for constructing the synthetic LLM questions. 
Our approach was to generate questions similar to those asked by real users by prompting the LLM to output: (1) a question based on paragraphs retrieved from the corpus, and (2) a "more general" version of the first question. We only use the "more general" set since they were more similar to real user queries.

\section{Embedding Models for Retrieval}
\label{sec:app_embedding_models}
We experimented with multiple top embedding models from the MTEB leader board to optimize retrieval for our system. These are outlined in \autoref{tab:embedding_models}.

\begin{table}[h]
\centering
\footnotesize
\begin{tabular}{|l|}
\hline
\textbf{HuggingFace embedding model name} \\ \hline
\texttt{Snowflake/snowflake-arctic-embed-m}\footnote{https://huggingface.co/Snowflake/snowflake-arctic-embed-m} \\ \hline
\texttt{sentence-transformers/all-mpnet-base-v2}\\ \citep{reimers-2019-sentence-bert} \\ \hline
\texttt{avsolatorio/GIST-Embedding-v0} \citep{Solatorio2024GISTEmbedGI} \\ \hline
\texttt{Snowflake/snowflake-arctic-embed-m-long} \footnote{https://huggingface.co/Snowflake/snowflake-arctic-embed-m-long} \\ \hline
\texttt{intfloat/e5-base-v2} \citep{wang2022text} \\ \hline
\texttt{mixedbread-ai/mxbai-embed-large-v1} \\\citep{emb2024mxbai} \\ \hline
\texttt{jinaai/jina-embeddings-v3} \citep{sturua2024jinaembeddingsv3multilingualembeddingstask} \\ \hline
\end{tabular}
\caption{Embedding Models to optimize retrieval}
\label{tab:embedding_models}
\end{table}

\section{Retrieval Ensemble Experiments}
\label{sec:app_ensemble}
\begin{figure}[ht]
    \centering
\includegraphics[width=1.0\linewidth]{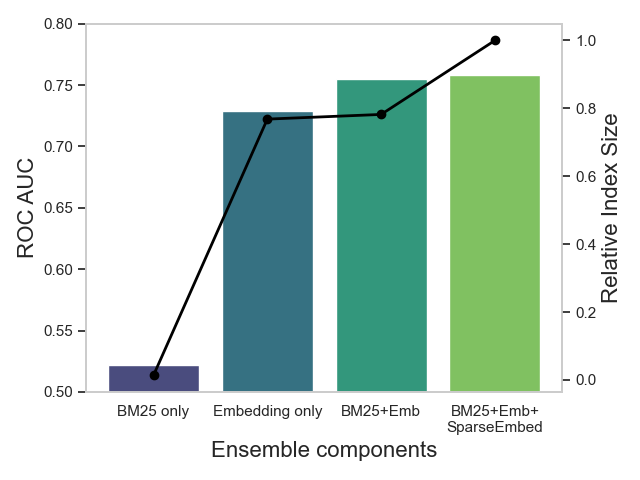}
    \caption{Ranking performance for various ensembles with relative size of the index required. Excluding SparseEmbed reduces the index size by 20\% without a significant drop in ranking performance.} 
    \label{fig:ranking-ensemble}
\end{figure}

\autoref{fig:ranking-ensemble} shows results of our ensembling experiments for the full-text retrieval index. SparseEmbed introduces an  overhead with minimal performance gains, so we picked an ensemble of embedding similarity and BM25 as our final ranking metric.

\section{Prompt for Evaluating Attribution}
\label{sec:app_att_prompt}
\begin{displayquote}
\footnotesize
\texttt{As an Attribution Validator, your task is to verify whether a given reference can support the given claim. 
A claim can be either a plain sentence or a question followed by its answer. 
Specifically, your response should clearly indicate the relationship: Attributable, Contradictory or Extrapolatory. 
A contradictory error occurs when you can infer that the answer contradicts the fact presented in the context, while an extrapolatory error means that you cannot infer the correctness of the answer based on the information provided in the context. 
Output your response as a json with only a single key "output" and a value of one among - ("Attributable", "Contradictory", "Extrapolatory").\\
Claim: {claim}\\
Reference: {ref\_excerpt}}
\end{displayquote}
\section{User Feedback Examples}
\label{sec:app_feedback}
 Table \ref{tab:feedback} lists some examples of the user complaints for Scholar QA reports.
\begin{table}[h]
\footnotesize
\centering
 \begin{tabular}{|p{0.97\linewidth}|}
\hline
\textbf{Feedback} \\ \hline
The structure is good, but the articles you choose are not from top journals. \\ \hline
The first citation says that \textit{rabbits} can obtain cholesterol from diet, not rats. \\ \hline
These provide a lot of general information about the topic, but nothing here actually addresses the central question I asked. \\ \hline
The answer did not address the `MOBILIZATION' techniques at all! The answer is wrong because it addressed Exercise therapy! \\ \hline
They address the general setting, but not the specific question I asked. \\ \hline
It's only analysing on SASAF model, but there are more. \\ \hline
\end{tabular}
\caption{Example Feedback on Research Issues}
\label{tab:feedback}
\end{table}

\section{Progress Updates and Report Sections}
\label{sec: app_progress}
\begin{figure}[h!]
    \centering
    \includegraphics[width=1\linewidth]{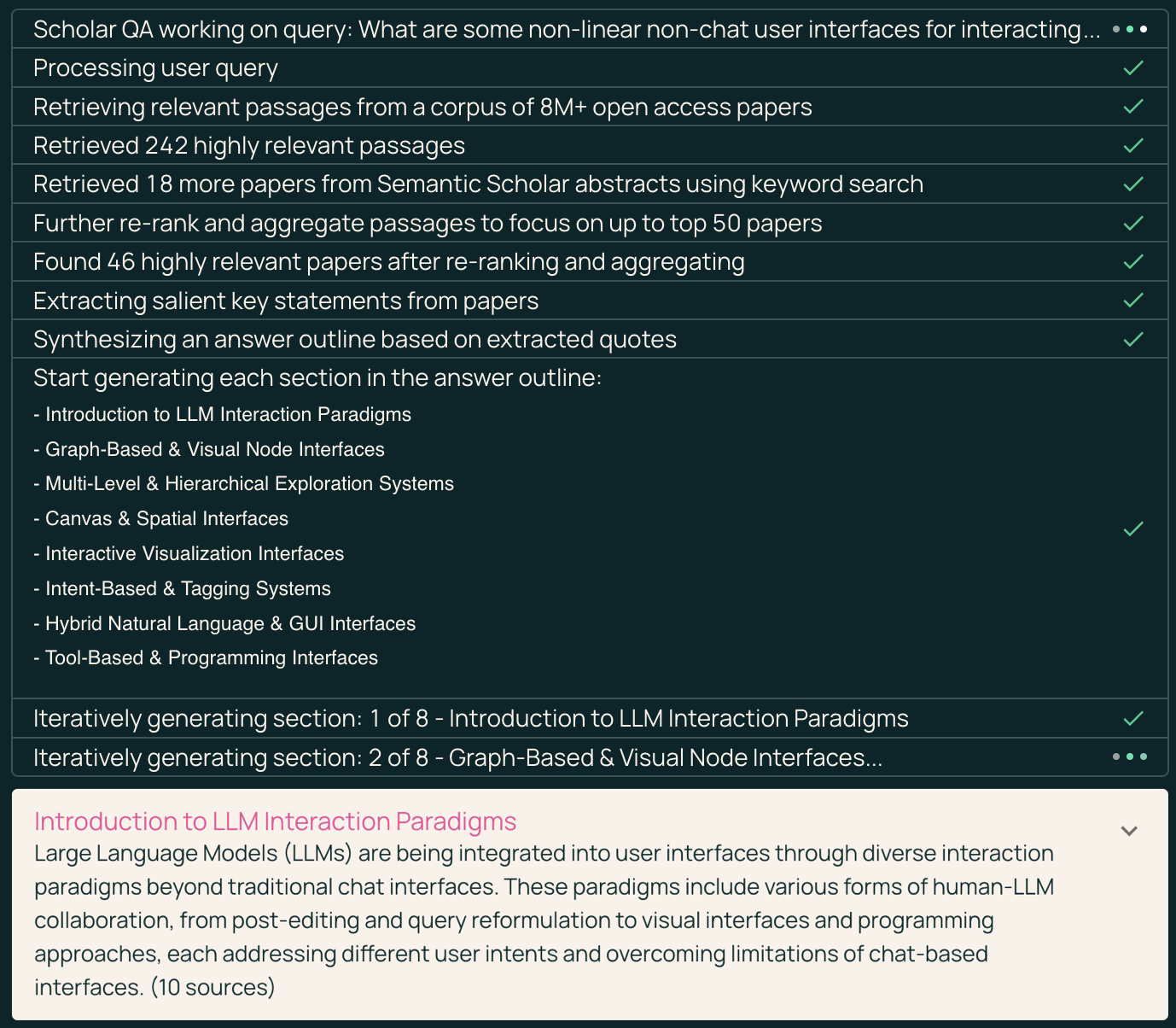}
    \caption{Progress indication and section streaming.}
    \label{fig:progress}
\end{figure}

Figure~\ref{fig:progress} demonstrates how we display in real-time the progress of the system during generation. This included number of papers and passages the were processed in each step, as well as the outline as it is being generated. Each section appears as soon as it is generated, so users can begin browsing the first sections.

\section{Query Type Analysis}\label{sec:app_query_types}
To analyze the types of questions users are asking, we use an LLM to categorize the queries. The most prominent types were comprehensive \emph{deep-dive} into a specific research topic (15k) and \emph{comparative analysis} of prior work (5k). Other themes such as factoid QA or specific methods, datasets accounted for fewer queries.
\begin{figure}[t!]
    \centering
    \includegraphics[width=1.1\linewidth]{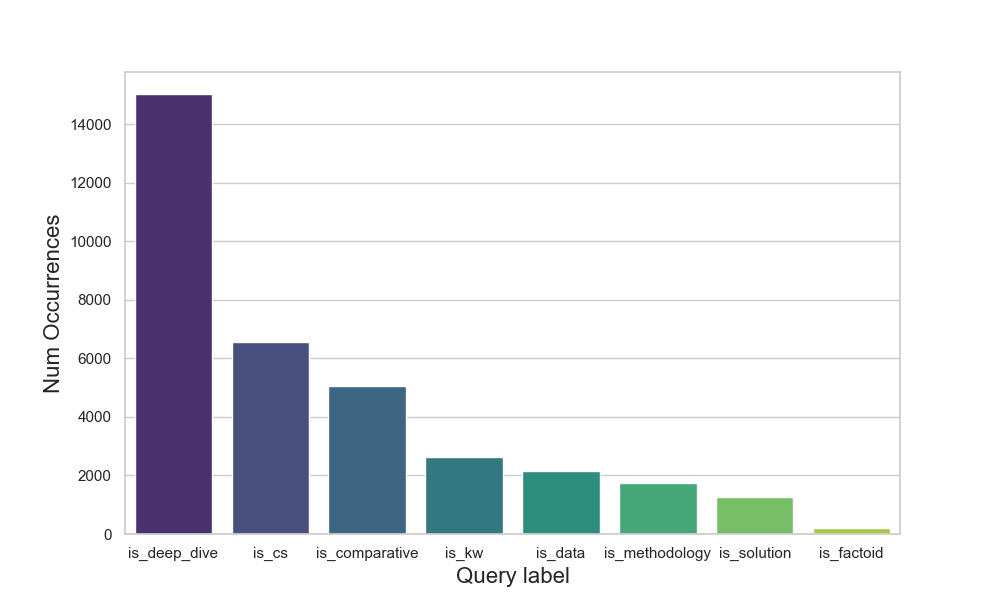}
    \caption{Distribution of different question types submitted to Scholar QA deployed Web application.}
    \label{fig:query-label}
\end{figure}

\section{Generation Results with updated GPT-4o}
\label{sec:app_gpt4o_updated_res}
\autoref{tab:sqa_eval_updated} shows results on ScholarQA-CS with {\tt gpt-4o-2024-11-20} as the LLM judge. These results can be contrasted with the first two columns in \autoref{tab:sqa_eval} which are obtained with {\tt gpt-4o-2024-08-06} as the judge. Even though the absolute scores are inflated compared to \autoref{tab:sqa_eval}, the relative rankings are about the same with Scholar QA getting the best overall score.
\begin{table}[ht]
\scriptsize
\centering
\setlength{\tabcolsep}{3pt}
\renewcommand{\arraystretch}{1}
\begin{tabular}{@{}lrr|lrrr@{}}
\toprule
\multirow{2}{*}{Model}   & \multicolumn{2}{c|}{Score}  & \multirow{2}{*}{Model}  & \multicolumn{2}{c}{Score}                                 \\ \cmidrule(l){2-3} \cmidrule(l){5-6} 
 & Rubrics& Total   & &Rubrics& Total  \\ \midrule
\multicolumn{3}{c|}{\emph{LLM Prompting (No Retrieval)}}     & \multicolumn{3}{c}{\emph{QA Systems}}               \\ \midrule

Llama 3.1-8B               &    51.8    &   48.2   &    
SQA-Claude 3.7 S           & {\ul 67.3} & {\ul \textbf{67.2}} & \\
Llama 3.1-70B              &    57.0    &   51.2     &  
SQA-Claude 3.5 S           & 61.3      & 67.1       &  \\

Claude 3.5 S               & 57.8       & 51.3      & 
OS-GPT-4o                  & 54.9       & 59.9       &  \\

Claude 3.7 S               & 68.4       & 60.8      &      
PaperQA2                   & 43.8       & 54.1      &  \\

\quad +Thinking  & 68.3  & 58.7      &
Perplex. Sonar DR          & 43.9       & 56.0      &  \\

GPT-4.1          & {\ul \textbf{69.3}}                & {\ul 61.8}     &  STORM            & 59.2                 & 64.7      & \\
o1-mini          & 69.1                & 61.3&   \\
o3-mini          & 68.5                 & 55.9    &   \\ \bottomrule

\end{tabular}
\caption{Evaluation results on ScholarQA-CS benchmark with {\tt gpt-4o-2024-11-20} as the judge. System responses are either generated by simply prompting LLMs with the questions or by issuing the queries to RAG based QA systems. Expert annotated rubrics only scores are reported in addition to the overall total. The overall best results are \textbf{highlighted} and best results within a category are {\ul underlined}. SQA: Ai2 Scholar QA, OS: Open Scholar, S: Sonnet, Claude 3.5 S: {\tt claude-3-5-sonnet-20241022}.}
\label{tab:sqa_eval_updated}
\vspace{-1.5em}
\end{table}

\end{document}